\documentclass[10pt,twocolumn,letterpaper]{article}

\usepackage[pagenumbers]{iccv}  
\usepackage{graphicx}          
\usepackage{amsmath}
\usepackage{multirow}           

\definecolor{iccvblue}{rgb}{0.21,0.49,0.74}
\usepackage[colorlinks=true,      
            allcolors=iccvblue,
            pdfborder={0 0 0},      
            pdftitle={REPT: A Robust Evaluation Platform for Transferability},
            pdfauthor={Helia Rezvani\textsuperscript{1,*} \qquad Alireza Kazemi\textsuperscript{1,*} \qquad Mahsa Baktashmotlagh\textsuperscript{1}}]{hyperref}

\title{Benchmarking Transferability: A Framework for Fair and Robust Evaluation}

\author{
    Alireza Kazemi\textsuperscript{1,*} \qquad Helia Rezvani\textsuperscript{1,*} \qquad Mahsa Baktashmotlagh\textsuperscript{1} \\
    {\small \texttt{a.kazemi@uq.edu.au} \qquad \texttt{h.rezvani@uq.edu.au} \qquad \texttt{m.baktashmotlagh@uq.edu.au}} \\
    \textsuperscript{1}The University of Queensland, Brisbane, Australia \\
    {\small * Equal contribution}
}

\begin{document}

\maketitle

\begin{abstract}

Transferability scores aim to quantify how well a model trained on one domain generalizes to a target domain. Despite numerous methods proposed for measuring transferability, their reliability and practical usefulness remain inconclusive, often due to differing experimental setups, datasets, and assumptions. In this paper, we introduce a comprehensive benchmarking framework designed to systematically evaluate transferability scores across diverse settings. Through extensive experiments, we observe variations in how different metrics perform under various scenarios, suggesting that current evaluation practices may not fully capture each method's strengths and limitations. Our findings underscore the value of standardized assessment protocols, paving the way for more reliable transferability measures and better-informed model selection in cross-domain applications. Additionally, we achieved a 3.5\% improvement using our proposed metric for the head-training fine-tuning experimental setup. Our code is available in this repository: \url{https://github.com/alizkzm/pert_robust_platform}.

\end{abstract}
\section{Introduction}
\label{sec:intro}

Transfer learning has emerged as a key approach in machine learning, largely due to increasing availability of pre-trained deep neural networks (DNNs)~\cite{Dwivedi2019}. At its core, transfer learning leverages knowledge gained from a source task to improve the performance on a new, target task~\cite{Ding2022, Tan2024}, making it particularly beneficial when labeled data for the target task is scarce or expensive to obtain.

The conventional deep learning paradigm often employs a model pre-trained on a large dataset as the initialization for a new task, adapting the model to the target dataset by fine-tuning some or all of its parameters~\cite{Dwivedi2019}. This strategy is particularly effective when training data are limited, since pre-trained weights can help prevent overfitting that might arise from training a deep model from scratch. Models pre-trained on datasets like ImageNet often serve as generic image representations applicable to a wide variety of recognition tasks~\cite{Dwivedi2020}, although the success of transfer learning can vary depending on the choice of source model. Selecting an appropriate source model can yield substantial improvements in accuracy, while a poorly matched model might underperform compared to training from scratch \cite{Tan2024}.

To address the challenge of source model selection~\cite{Tan2021}, transferability estimation metrics have been proposed that estimate how well a pre-trained model might perform on a target task without incurring the full cost of fine-tuning~\cite{Tan2021}. These metrics provide a score that ideally correlates with real transfer performance after fine-tuning, enabling practitioners to efficiently identify the most promising models for transfer learning. Recent studies have proposed vairous transferability estimation metrics, such as LEEP \cite{Nguyen2020}, LogME \cite{You2021}, PACTran \cite{Ding2022}, SFDA \cite{Shao2022}, and ETran \cite{Gholami2023}, that evaluate a pre-trained model’s suitability for a target task by analyzing extracted features, thereby bypassing the costly process of fine-tuning multiple models. However, despite their success, existing approaches face several limitations, listed below.

\noindent\textbf{Dependency on Target Labels:} Many existing transferability metrics, including state-of-the art techniques of LEEP~\cite{Nguyen2020}, LogME~\cite{You2021}, SFDA~\cite{Shao2022}, and ETran~\cite{Gholami2023} (which is partially dependent on labels), depend on labeled target data for estimating transferability, which limits their applicability when labels are scarce, noisy, or unavailable.

\noindent\textbf{Source Dataset Assumptions:} Most transferability metrics assume that all source models are pre-trained on a large-scale dataset like ImageNet. This assumption limits their practical applicability, as access to ImageNet-pretrained models is not always feasible. In many real-world scenarios, we may only have access to models trained on domain-specific or smaller-scale datasets. Additionally, recent research~\cite{Fang2024} shows that improvements in ImageNet accuracy do not consistently translate to better performance on real-world datasets. As a result, these metrics can become unreliable when the source distribution deviates from the ImageNet assumption.

\noindent\textbf{Model Complexity Considerations:} Previous methods implicitly assume that model hubs contain only state-of-the-art, high-performance architectures. However, real-world applications often require a trade-off between model complexity and computational efficiency. While transferability scores are typically designed for more complex models, many practical scenarios involve lightweight or compressed models optimized for efficiency \cite{Xu2024}. These models may not exhibit the same feature representations as their high-performance counterparts, leading to inconsistencies in transferability assessment when applied to models of varying complexities.

\noindent\textbf{Fine-tuning Strategy Variations:} Most existing transferability metrics focus primarily on vanilla fine-tuning, where all parameters of the pre-trained model are updated. However, in practice, various fine-tuning strategies are employed, such as last-layer tuning or feature extraction with a new classifier. Current metrics often fail to account for different adaptation approaches, potentially leading to suboptimal model selection for specific fine-tuning strategies.

In this work, we systematically analyze how sensitive existing transferability metrics are to varying problem setups, which underscores the critical need for designing metrics that are robust and consistently reliable. Our contributions are three-folded:
\begin{itemize}
    \item \textbf{A Systematic Evaluation Framework:} We introduce a platform that enables evaluating transferability of learning models under various problem settings, including different source datasets, model complexities, fine-tuning strategies, and label availability.
    
    \item \textbf{Label-Free Transferability Estimation:} We propose a simple, yet effective approach for transferability estimation that does not depend on target labels, making it applicable in scenarios where labeled data is scarce or unavailable. Our method achieves comparable performance to label-dependent metrics while demonstrating more stable results across different problem settings.
    
    
    \item \textbf{Source Dataset Independence:} Our approach shows robust performance even when source models are trained on different datasets, addressing a key limitation of existing methods that assume ImageNet pre-training.
\end{itemize}

\section{Related Work}

\subsection{Transferability Estimation Metrics}

The challenge of efficiently selecting optimal pre-trained models has driven the development of various transferability estimation metrics that avoid the computational expense of exhaustive fine-tuning.

Early approaches focused on comparing source and target label spaces. Negative Conditional Entropy (NCE) \cite{Tran2019} measures the conditional entropy between source and target label distributions, providing an early framework for transferability estimation. Building on this concept, Log Expected Empirical Predictor (LEEP) \cite{Nguyen2020} improved transferability estimation by computing the expected empirical conditional distribution between source predictions and target labels, estimating the joint probability over source and target label spaces.

To overcome the limitation of requiring source model classifiers, feature-based methods emerged. N-LEEP \cite{Li2021} extended LEEP by replacing source classifiers with Gaussian Mixture Models to handle self-supervised models. LogME \cite{You2021} proposed a more principled approach by estimating the maximum evidence (marginalized likelihood) of target labels given extracted features. It models the relationship between features and labels using a Bayesian framework, where evidence is calculated by integrating over all possible values of model weights rather than using a single optimal value, making it more robust to overfitting than maximum likelihood methods.

Recent methods have focused on class separability as a key indicator of transferability potential. SFDA \cite{Shao2022} employs Fisher Discriminant Analysis to project features into more discriminative spaces through a self-challenging mechanism. It first embeds static features into a Fisher space for better separability, then applies a "confidence mixing" noise that increases classification difficulty, encouraging models to differentiate on hard examples. This two-stage approach better simulates the dynamics of fine-tuning compared to static feature evaluation.

Energy-based approaches have also emerged, with ETran \cite{Gholami2023} introducing a framework combining energy, classification, and regression scores. ETran uses energy-based models to detect whether a target dataset is in-distribution or out-of-distribution for a given pre-trained model. The energy score evaluates the likelihood of features being in-distribution data for the pre-trained model, while classification scores project features to discriminative spaces using Linear Discriminant Analysis, and regression scores utilize Singular Value Decomposition to efficiently estimate transferability. This comprehensive approach makes ETran applicable to classification, regression, and even object detection tasks, which previous metrics could not address.

PACTran \cite{Ding2022} provides a theoretical foundation through PAC-Bayesian theory, establishing guarantees on transferability estimation, while NCTI \cite{Wang2023} leverages neural collapse theory to measure the distance between current feature representations and their hypothetical post-fine-tuning state.

\subsection{Challenges in Transferability Estimation}

Despite these advances, current transferability estimation research faces several important limitations that hinder their practical applicability. Most existing transferability metrics require labeled target data for their estimations, making them less useful in scenarios where annotations are scarce or expensive to obtain. Current methods also typically assume all candidate models are pre-trained on the same source dataset (commonly ImageNet), despite evidence from \citet{Fang2024} showing that improvements in ImageNet performance don't necessarily translate to better downstream task performance. Additionally, existing evaluation frameworks exhibit a bias toward high-capacity models, neglecting the growing importance of efficient, lightweight architectures that are essential for resource-constrained applications \cite{Xu2024}. Furthermore, most transferability studies focus solely on vanilla fine-tuning, overlooking alternative strategies such as feature extraction with linear classifiers or partial fine-tuning that practitioners frequently employ. These limitations collectively restrict the effectiveness of existing transferability metrics in realistic deployment scenarios, where diverse pre-training sources, model architectures, and fine-tuning approaches are common. Our work addresses these challenges through a comprehensive evaluation framework and a label-free transferability estimation approach that maintains effectiveness across diverse scenarios.
\section{Robust Transferability Evaluation}

\subsubsection{Preliminaries}

Let $T = \{X, Y\}$ represent the target dataset, and $\{\phi_l\}_{l=1}^L$ denote $L$ pre-trained feature-extractors, with $l$ being the index. Using transferability estimation, given by $M$, our goal is to rank these pre-trained models based on their performance on a given target dataset. We assess the transferability of each pre-trained model using a specific metric $M$ that produces a numerical score, denoted as $T_l$. Essentially, a higher score for $T_l$ implies that the model $\phi_l$ is more likely to perform effectively on the given target dataset.


\noindent\textbf{Source-Hub}: For pre-training our models, we utilize multiple diverse datasets including Caltech101~\cite{Li2022}, Caltech256, CIFAR-100~\cite{Krizhevsky2009}, Flowers102~\cite{Nilsback2008}, and ImageNet. This diversity allows us to comprehensively evaluate transferability across varying source domains.

\noindent\textbf{Target-Hub}: We consider a wide range of classification benchmark datasets, including five fine-grained classifications (FGVC Aircraft~\cite{Maji2013}, Stanford Cars~\cite{Krause2013}, Food101~\cite{Bossard2014}, Oxford-IIIT Pets~\cite{Parkhi2012}, Oxford-102 Flowers~\cite{Nilsback2008}), five coarse-grained classifications (Caltech256, Caltech101~\cite{Li2022}, CIFAR-10~\cite{Krizhevsky2009}, CIFAR-100~\cite{Krizhevsky2009}, VOC2007~\cite{Everingham2015}), one scene classification (SUN397), and one texture classification (DTD~\cite{Cimpoi2014}). In total, we adopted 12 benchmark datasets for use in sources and targets, providing a comprehensive evaluation landscape across varying visual domains.


\noindent\textbf{Scores}: We evaluate both state-of-the-art transfer learning metrics and our proposed label-free baselines. The state-of-the-art metrics include ETran~\cite{Gholami2023}, SFDA~\cite{Shao2022}, LogME~\cite{You2021}, and PACTran~\cite{Ding2022}. Additionally, we implement distribution-based metric of Wasserstein. In this section, we elaborate on these metrics and introduce our novel weight-based approach for transferability estimation.


\noindent\textbf{Evaluation Metric}: To evaluate the performance of transferability metrics, we need ground-truth ranking scores of all pre-trained models ($\Phi_m$). These ground-truth scores, denoted by $G_m$, are the validation accuracies obtained after fine-tuning each $\Phi_m$ on the target dataset. Following previous works~\cite{Shao2022, You2021, Nguyen2020}, we use Kendall's tau~\cite{Vigna2015}, denoted by $\tau$, as our main evaluation metric. Kendall's tau is defined as the number of concordant pairs minus the number of discordant pairs divided by the overall number of pairs:

\begin{equation}
\tau = \frac{2}{M(M-1)} \sum_{i=1}^M \sum_{j=i+1}^M \text{sgn}(G_i - G_j) \cdot \text{sgn}(T_i - T_j)
\end{equation}

where $\text{sgn}$ is the sign function, $G_i$ and $G_j$ are the ground-truth performances, and $T_i$ and $T_j$ are the estimated transferability scores for models $i$ and $j$, respectively. A higher value of $\tau$ indicates a stronger correlation between the estimated transferability ranking and the actual performance after fine-tuning.

\noindent\textbf{Wasserstein Distance:} The Wasserstein distance (also known as Earth Mover's Distance) provides a natural way to compare probability distributions by measuring the minimum "work" required to transform one distribution into another~\cite{Villani2009, Peyre2019}. In the context of transferability estimation, we use the Wasserstein distance to measure the discrepancy between the weight distributions of the original pre-trained model and the model after fine-tuning on pseudo-labeled target data.

Given two probability distributions $P$ and $Q$, the 1-Wasserstein distance is defined as:

\begin{equation}
W_1(P, Q) = \inf_{\gamma \in \Gamma(P, Q)} \mathbb{E}_{(x,y) \sim \gamma} [\|x - y\|]
\end{equation}

where $\Gamma(P, Q)$ denotes the set of all joint distributions $\gamma(x, y)$ whose marginals are $P$ and $Q$.

In our transferability estimation context, we calculate the Wasserstein distance between the distribution of the original model parameters $\theta_l$ and the fine-tuned parameters $\theta_l'$:

\begin{equation}
T_l^{W} = -W_1(\mathcal{P}_{\theta_l}, \mathcal{P}_{\theta_l'})
\end{equation}

where $\mathcal{P}_{\theta_l}$ and $\mathcal{P}_{\theta_l'}$ represent the distributions of the parameters before and after fine-tuning, respectively. The Wasserstein distance provides several advantages over simpler metrics like L1 and L2:

\begin{itemize}
    \item It accounts for the overall distribution shift rather than just element-wise differences
    \item It is particularly effective for comparing distributions with different supports
    \item It provides a more geometrically meaningful comparison between weight distributions
\end{itemize}

These properties make Wasserstein distance especially effective for transferability estimation across different model architectures and training regimes, as our experiments demonstrate.

\subsubsection{Weight-Based Transferability Metric}

At the core of our extensive transferability estimation experiments is a PyTorch-based platform designed to facilitate reproducible and systematic research in transfer learning. The platform enables the evaluation of various transferability scores under different settings by controlling the source dataset, model complexity, and fine-tuning strategy. The repository is structured to allow easy experimentation with new models, datasets, and scoring techniques, making it a dynamic and extensible framework. 

Unlike existing methods that rely heavily on target labels, our proposed approach aims to provide reliable transferability estimates without requiring labeled target data, making it more applicable to real-world scenarios where labels may be scarce or unavailable. Our approach draws inspiration from Projection Norm~\cite{Yu2022}, which was originally designed to predict a model's performance on out-of-distribution (OOD) data without access to ground truth labels. The key insight is that models with greater adaptability to new data distributions show particular patterns in weight changes during fine-tuning on that data.

We assess transferability through a three-step process:

\begin{enumerate}
\item Using the pre-trained model to generate pseudo-labels for the target dataset
\item Fine-tuning the model for a short period (two epochs) on these pseudo-labeled samples
\item Measuring Wasserstein distance metric between the original model weights and the fine-tuned weights
\end{enumerate}

Formally, given a pre-trained model $\phi_l$ with parameters $\theta_l$, we generate pseudo-labels $\hat{Y}$ for the target dataset $X$:

\begin{equation}
\hat{Y} = \arg\max \phi_l(X)
\end{equation}

We then fine-tune the model for two epochs on $(X, \hat{Y})$ to obtain updated parameters $\theta_l'$. The transferability score is calculated using the Wasserstein distance metric between $\theta_l$ and $\theta_l'$:

\begin{equation}
T_l^{\text{Wasserstein}} = -W(\theta_l, \theta_l')
\end{equation}

The negative sign ensures that a smaller distance (indicating better transferability) corresponds to a higher score. This approach provides a label-free method for transferability estimation that focuses on how easily the model adapts to the target distribution, bypassing the need for ground truth labels that most existing methods require.

\subsection{Evaluation Criteria}

To systematically analyze transferability scores, TransferTest allows users to configure experiments based on three key implementation choices:

\begin{enumerate}
\item \textbf{Source Dataset Selection}: Users can specify which pre-trained models to use based on their source datasets. This enables controlled experiments on how different pre-training data distributions influence transferability, a critical factor often overlooked in previous research that typically assumes ImageNet pre-training.

\item \textbf{Model-Hub Complexity}: The platform allows filtering models based on their complexity, from shallow to deeper models~\cite{He2016, Huang2017, Sandler2018, Szegedy2015, Szegedy2016}. This helps evaluate whether more complex models provide better transferability or if lightweight architectures are sufficient, addressing practical deployment constraints.

\item \textbf{Fine-Tuning Strategy}: It allows selecting head-training over full-training to understand how different adaptation strategies affect transferability metrics~\cite{You2021, Shao2022, Wang2023}. This feature acknowledges that different fine-tuning approaches may be appropriate in different scenarios depending on computational resources and dataset characteristics.
\end{enumerate}

By systematically varying these three factors, TransferTest provides a structured approach to evaluating transferability scores under diverse scenarios, ensuring reliable and reproducible assessments that more accurately reflect real-world transfer learning challenges.

\section{Experiments}

In this section, we present comprehensive experiments evaluating transferability metrics across different source datasets, model complexities, and fine-tuning strategies. Our analysis demonstrates the advantages of our proposed weight-based approaches, particularly in challenging scenarios where source datasets differ from ImageNet or when working with limited model pools.

\subsection{Experimental Setup}

\textbf{Model-Hub}: We evaluate transferability estimation metrics on two model pools:

1) \textbf{Supervised Models}: We use 11 widely adopted architectures pre-trained on different source datasets: ResNet-34~\cite{He2016}, ResNet-50~\cite{He2016}, ResNet-101~\cite{He2016}, ResNet-151~\cite{He2016}, DenseNet-121~\cite{Huang2017}, DenseNet-169~\cite{Huang2017}, DenseNet-201~\cite{Huang2017}, MNet-A1~\cite{Tan2019}, MobileNet-v2~\cite{Sandler2018}, GoogleNet~\cite{Szegedy2015}, and Inception-v3~\cite{Szegedy2016}.

2) \textbf{Self-Supervised Models}: We construct a pool with 10 SSL pre-trained models, including MoCo-v1~\cite{He2020}, MoCo-v2~\cite{Chen2020}, PCL-v2~\cite{Li2020}, SELA-V2, Deepcluster-v2~\cite{Caron2020}, BYOL~\cite{Grill2020}, Infomin~\cite{Tian2020}, SWAV~\cite{Caron2018}, and InstDis~\cite{Wu2018}.

All models with source data other than ImageNet are trained with standardized configurations to achieve state-of-the-art performance on their respective benchmarks.

\noindent\textbf{Evaluation Metric}: Following previous work~\cite{Shao2022, You2021}, we use weighted Kendall's tau correlation~\cite{Vigna2015} to measure how well transferability metrics predict the actual ranking of models after fine-tuning. This metric assigns higher importance to correctly ranking top-performing models, better reflecting practical model selection scenarios.
\subsubsection{Implementation Details}

For our experiments, we utilize pre-trained source models available in PyTorch~\cite{paszke2019pytorch} and additionally train our model hub on other datasets including CIFAR-100~\cite{Krizhevsky2009}, Caltech101~\cite{Li2022}, Caltech256, and Flowers102~\cite{Nilsback2008}. These models were trained for 200 epochs with early stopping and hyperparameter optimization. For fine-tuning, we follow a grid search approach, selecting learning rates from $\{10^{-1}, 10^{-2}, 10^{-3}, 10^{-4}\}$ and weight decay parameters from $\{10^{-6}, 10^{-5}, 10^{-4}, 10^{-3}\}$ as described in~\cite{You2021}. These fine-tuned models achieved state-of-the-art performance for each source model.

For ground truth measurements, we consider two different fine-tuning strategies. For full-training, we adopt the benchmarks from~\cite{Shao2022}. For head-training, we employ the same setup as in full-training but limit the process to two epochs.

\subsection{Experimental Results}

\subsubsection{Performance Across Source Datasets}

Most existing transferability metrics have been evaluated primarily on models pre-trained on ImageNet. However, this approach fails to reflect many practical scenarios where models are trained on domain-specific or proprietary datasets. We evaluate how transferability estimation metrics perform when the source dataset varies.

\begin{table*}[t]
    \centering
    \caption{The impact of source datasets on transferability estimation: This table shows weighted Kendall's tau correlations using ETran score across different source datasets for supervised models. Higher values indicate better transferability estimation. Results reveal that while ImageNet provides strong overall performance, other source datasets may be better for specific target domains (highlighted in bold).}
    \label{tab:data_source_top_1}
    \resizebox{\textwidth}{!}{ 
    \begin{tabular}{lccccccccccc|c}
        \toprule
        Source & Aircraft & Caltech101 & Cars & Cifar10 & Cifar100 & DTD & Flowers & Food & Pets & SUN & VOC & Average\\
        \midrule
        ImageNet & -0.091 & 0.440 & 0.246 & \textbf{0.887} & \textbf{0.900} & 0.303 & 0.580 & 0.829 & \textbf{0.713} & \textbf{0.708} & \textbf{0.667} & 0.562 
        \\
        CIFAR-100~\cite{Krizhevsky2009} & -0.205 & 0.321 & 0.119 & 0.363 & - & -0.078 & 0.039 & 0.298 & 0.133 & 0.549 & -0.109 & 0.143 \\
        Caltech101~\cite{Li2022} & 0.105 & - & 0.269 & 0.347 &0.397 & 0.194 & 0.261 & \textbf{0.431} & -0.003 & 0.687 & -0.233 & 0.245 \\
        Caltech256 & \textbf{0.219} & \textbf{0.562} & \textbf{0.321} & 0.432 & 0.488 & 0.337 & \textbf{0.653} & 0.388 & 0.421 & 0.587 & 0.265 & 0.425 \\
        Flowers102~\cite{Nilsback2008} & -0.028 & -0.104 & 0.128 & 0.541 & 0.544 & \textbf{0.378} & - & 0.421 & 0.434 & 0.576 & 0.373 & 0.326\\
        \bottomrule
    \end{tabular}
    }
\end{table*}

\begin{table*}[!ht]
    \centering
    \caption{SFDA performance across source datasets: This table presents weighted Kendall's tau correlations using SFDA across different source datasets. Bold values indicate best performance for each target dataset. Note the significant performance degradation when moving away from ImageNet pre-training for most datasets, highlighting the source dataset dependency of existing methods.}
    \label{tab:data_source_top3}
    \resizebox{\textwidth}{!}{ 
    \begin{tabular}{lccccccccccc|c}
        \toprule
        Source & Aircraft & Caltech101 & Cars & Cifar10 & Cifar100 & DTD & Flowers & Food & Pets & SUN & VOC & Average\\
        \midrule
        ImageNet & -0.215 & \textbf{0.521} & \textbf{0.462} & \textbf{0.849} & \textbf{0.793} & 0.633 & 0.590 & \textbf{0.478} & 0.341 & \textbf{0.532} & 0.518 & 0.501  \\
        CIFAR-100~\cite{Krizhevsky2009} & -0.008 & 0.291 & 0.254 & 0.408 & - & 0.361 & 0.546 & 0.273 & 0.204 & 0.312 & 0.240 & 0.292  \\
        Caltech101~\cite{Li2022} & \textbf{0.367} & - & 0.251 & 0.214 & 0.402 & 0.562 & \textbf{0.711} & 0.181 & 0.160 & 0.131 & 0.252 & 0.211  \\
        Caltech256 & -0.241 & 0.352 & 0.312 & 0.388 & 0.475 & 0.538 & 0.140 & 0.331 & 0.352 & 0.361 & \textbf{0.525} & 0.322  \\
        Flowers102~\cite{Nilsback2008} & 0.111 & 0.493 & 0.431 & 0.743 & 0.696 & \textbf{0.658} & - & 0.471 & \textbf{0.497} & 0.521 & 0.379 & 0.514  \\
        \bottomrule
    \end{tabular}
    }
\end{table*}

\begin{table*}[!ht]
    \centering
    \caption{LogME performance across source datasets: This table shows the weighted Kendall's tau correlations using LogME for different source-target combinations. The performance variance across datasets demonstrates that LogME's effectiveness is highly dependent on the source dataset used for pre-training.}
    \label{tab:data_source_logme}
    \resizebox{\textwidth}{!}{ 
    \begin{tabular}{lccccccccccc|c}
        \toprule
        Source & Aircraft & Caltech101 & Cars & Cifar10 & Cifar100 & DTD & Flowers & Food & Pets & SUN & VOC & Average\\
        \midrule
        ImageNet & \textbf{0.405} & \textbf{0.523} & \textbf{0.476} & \textbf{0.852} & \textbf{0.725} & 0.662 & -0.008 & \textbf{0.489} & 0.411 & \textbf{0.517} & \textbf{0.695} & 0.534  \\
        CIFAR-100~\cite{Krizhevsky2009} & 0.046 & 0.281 & 0.263 & 0.593 & - & 0.615 & 0.098 & 0.275 & 0.188 & 0.312 & 0.024 & 0.261  \\
        Caltech101~\cite{Li2022} & 0.008 & - & 0.234 & 0.365 & 0.395 & \textbf{0.671} & \textbf{0.106} & 0.187 & 0.107 & 0.214 & -0.151 & 0.214  \\
        Caltech256 & 0.285 & 0.359 & 0.330 & 0.535 & 0.659 & 0.566 & 0.092 & 0.348 & \textbf{0.426} & 0.389 & 0.130 & 0.385  \\
        Flowers102~\cite{Nilsback2008} & 0.240 & 0.419 & 0.401 & 0.498 & 0.568 & 0.597 & - & 0.393 & 0.203 & 0.321 & 0.265 & 0.395  \\
        \bottomrule
    \end{tabular}
    }
\end{table*}

Tables~\ref{tab:data_source_top_1}-\ref{tab:data_source_logme} show that the performance of existing transferability metrics drops significantly when models are pre-trained on datasets other than ImageNet. For instance, ETran's average correlation decreases from 0.562 with ImageNet to 0.143 with CIFAR-100. However, an interesting pattern emerges: for specific target datasets, non-ImageNet source datasets can provide better transferability estimates. For example, Caltech256-trained models yield better transferability estimates for Aircraft, Caltech101, Cars, and Flowers datasets.

Next, we evaluate our proposed Wasserstein distance-based metric across different source datasets:

\begin{table*}[!ht]
    \centering
    \caption{Wasserstein distance metric performance: This table demonstrates our proposed Wasserstein distance-based metric's performance across various source-target combinations. Unlike feature-based metrics, our approach shows more consistent performance across different source datasets, suggesting it's less reliant on specific pre-training distributions.}
    \label{tab:wassertian}
    \resizebox{\textwidth}{!}{ 
    \begin{tabular}{lccccccccccc|c}
        \toprule
        Source & Aircraft & Caltech101 & Cars & Cifar10 & Cifar100 & DTD & Flowers & Food & Pets & SUN & VOC & Average\\
        \midrule
        ImageNet & 0.363 & 0.382 & 0.287 & 0.301 & \textbf{0.550} & 0.334 & \textbf{0.715} & 0.538 & \textbf{0.585} & 0.283 & \textbf{0.697} & 0.458 \\
        CIFAR-100~\cite{Krizhevsky2009} & 0.201 & -0.074 & 0.343 & 0.152 & - & 0.124 & 0.330 & \textbf{0.593} & 0.111 & \textbf{0.771} & 0.541 & 0.327  \\
        Caltech101~\cite{Li2022} & 0.294 & \textbf{-0.274} & \textbf{0.549} & -0.214 & 0.451 & 0.219 & 0.060 & 0.510 & 0.315 & 0.321 & 0.530 & 0.251  \\
        Caltech256 & \textbf{0.541} & -0.101 & 0.450 & 0.299 & 0.454 & \textbf{0.358} & 0.315 & 0.217 & 0.456 & 0.458 & 0.486 & 0.358  \\
        Flowers102~\cite{Nilsback2008} & 0.015 & \textbf{0.921} & 0.279 & \textbf{0.537} & 0.539 & 0.303 & 0.101 & 0.443 & -0.222 & 0.211 & -0.010 & 0.283  \\
        \bottomrule
    \end{tabular}
    }
\end{table*}

\begin{table}[t]
\caption{Comparison of transferability metrics across different source datasets for supervised and self-supervised models. Bold values indicate the best metric for each source dataset. Our Wasserstein-based approach demonstrates robust performance across diverse source datasets, particularly excelling with non-ImageNet sources.}
\label{tab:comparing_metrics}
\begin{subtable}{\linewidth}
\caption{Supervised Models}
\label{tab:supervised_models}
\centering
\begin{tabular}{lccc|c}
\toprule
Source & ETran & SFDA & LogME & Wasserstein \\
\midrule
ImageNet & \textbf{0.562} & 0.501 & 0.534 & 0.458 \\ 
CIFAR-100~\cite{Krizhevsky2009} & 0.143 & 0.292 & 0.261 & \textbf{0.327} \\ 
Caltech101~\cite{Li2022} & 0.245 & 0.211 & 0.214 & \textbf{0.251} \\ 
Caltech256 & \textbf{0.425} & 0.322 & 0.385 & 0.358 \\ 
Flowers102~\cite{Nilsback2008} & 0.326 & \textbf{0.514} & 0.395 & 0.283\\ 
\bottomrule
\end{tabular}
\end{subtable}

\vspace{1em}

\begin{subtable}{\linewidth}
\caption{Self-supervised Models}
\label{tab:self_supervised_models}
\centering
\begin{tabular}{lcc|c}
\toprule
Source & SFDA & LogME & Wasserstein \\
\midrule
ImageNet & \textbf{0.685} & 0.411 & 0.451 \\ 
CIFAR-100~\cite{Krizhevsky2009} & 0.267 & 0.352 & \textbf{0.361} \\ 
Caltech101~\cite{Li2022} & \textbf{0.455} & 0.312 & 0.348 \\ 
Caltech256 & 0.409 & 0.218 & \textbf{0.437} \\ 
Flowers102~\cite{Nilsback2008} & 0.399 & \textbf{0.412} & 0.301\\ 
\bottomrule
\end{tabular}
\end{subtable}
\end{table}

\begin{figure}[!ht]
    \centering
    \begin{subfigure}{0.48\textwidth}
        \centering
        \includegraphics[width=\linewidth]{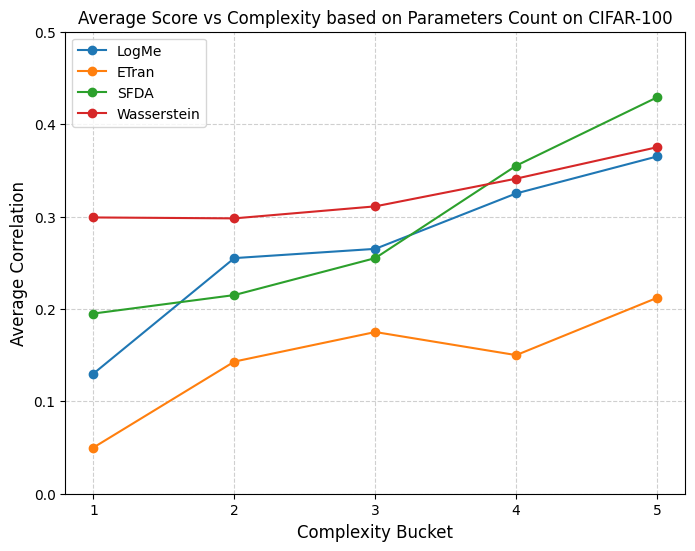}
        \caption{CIFAR-100 transferability estimation with varying model complexity}
    \end{subfigure}

    \begin{subfigure}{0.48\textwidth}
        \centering
        \includegraphics[width=\linewidth]{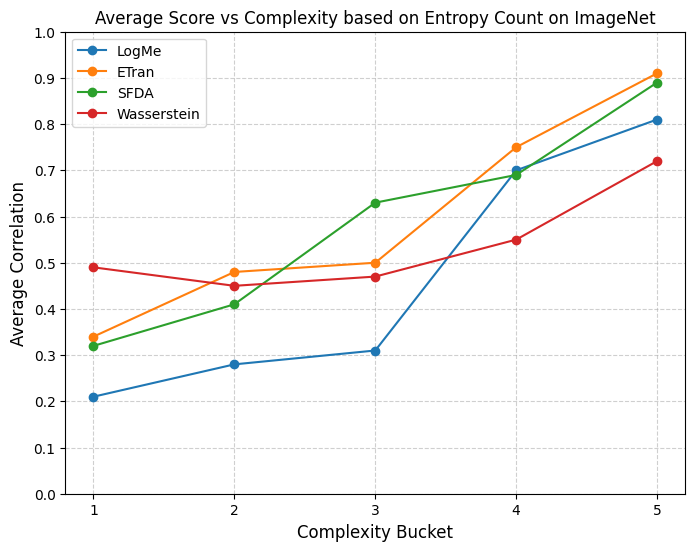}
        \caption{ImageNet transferability estimation with varying model complexity}
    \end{subfigure}

    \caption{Impact of model complexity on transferability metrics. The x-axis represents seven distinct model subsets, ordered by increasing parameter count and thus complexity (level 1: First models with the lowest parameters, level 5: Last models with the highest parameters, levels 2-4: intermediate complexity selections). The y-axis displays the weighted Kendall’s tau correlation between the predicted transferability rankings of different metrics and the actual transfer learning performance. Our proposed weight-based metrics (shown in green and purple) exhibit consistently higher correlation across varying model complexity levels compared to feature-based transferability estimation methods such as ETran, LogME, and SFDA.}
    \label{fig:model_complexity_figure}
\end{figure}

Table~\ref{tab:comparing_metrics} presents a comparison of different transferability metrics across source datasets. A key finding is that while ETran performs best with ImageNet pre-training, our Wasserstein distance metric outperforms other approaches for CIFAR-100 and Caltech101 source datasets in the supervised model hub (Table~\ref{tab:supervised_models}) and for CIFAR-100 and Caltech256 source datasets in the self-supervised model hub (Table~\ref{tab:self_supervised_models}). This highlights the robustness of our weight-based approach when applied to models trained on diverse source datasets, effectively addressing a key limitation of existing metrics.

\subsubsection{Impact of Model Complexity}

In real-world scenarios, practitioners often have access to a limited set of pre-trained models with varying architectures and complexities. We investigate how transferability metrics perform when the pool of available models is constrained by complexity.

Figure~\ref{fig:model_complexity_figure} reveals a clear trend across both CIFAR-100 and ImageNet experiments: as model complexity decreases (moving from right to left on the x-axis), the performance of all transferability metrics declines. However, our weight-based metrics demonstrate significantly greater stability compared to feature-based approaches like ETran~\cite{Gholami2023}, LogME~\cite{You2021}, and SFDA~\cite{Shao2022}. 

This resilience to model complexity variations is particularly valuable in resource-constrained environments, where only lightweight models might be available for transfer learning. The stability of our weight-based metrics suggests they capture more fundamental aspects of transferability that persist across model scales.
\begin{figure}[t]
    \centering
    \includegraphics[width=\linewidth]{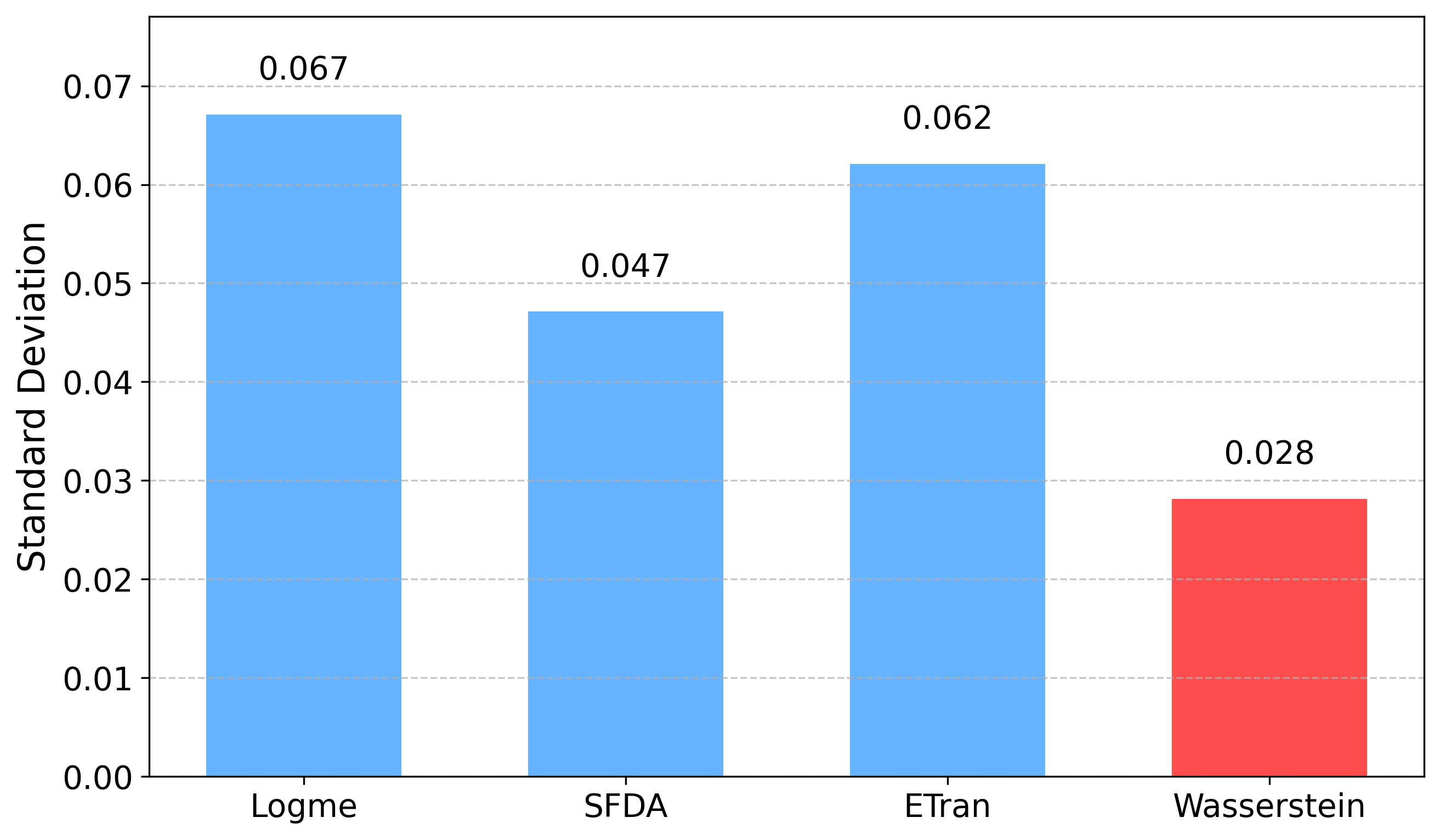}
    \caption{Comparison of standard deviations across different transferability metrics. The Wasserstein metric (our proposed method) demonstrates significantly lower variance ($\sigma = 0.028$) compared to alternative metrics when evaluated on different subsets of the supervised model-hub. Lower standard deviation indicates more consistent performance across diverse model architectures and datasets.}
    \label{fig:method_variance}
\end{figure}

\subsubsection{Effect of Fine-tuning Strategy}

The choice of fine-tuning strategy—whether to train only the classification head, the last few layers, or the entire network—can significantly impact both the transfer performance and the effectiveness of transferability estimation metrics~\cite{Dwivedi2019, You2021}. We evaluate how different metrics perform under varying fine-tuning approaches.

\begin{table}[t]
\centering
\caption{Performance comparison of head-training across different source datasets and supervised model-hub: This table shows the average correlation (weighted Kendall's tau) for different transferability metrics under head-training fine-tuning strategy across multiple source datasets.}
\label{tab:fine_tuning}

\begin{subtable}{\linewidth}
\centering
\caption{Source: ImageNet}
\label{tab:imagenet}
\resizebox{0.8\linewidth}{!}{
\begin{tabular}{lccc|c}
\toprule
Score & ETran & SFDA & LogME & Wasserstein \\
\midrule

Head-training & 0.431 & 0.453 & 0.501 & \textbf{0.523} \\
\bottomrule
\end{tabular}
}
\end{subtable}

\vspace{1em}
\begin{subtable}{\linewidth}
\centering
\caption{Source: CIFAR100}
\label{tab:cifar100}
\resizebox{0.8\linewidth}{!}{
\begin{tabular}{lccc|c}
\toprule
Score & ETran & SFDA & LogME & Wasserstein \\
\midrule
Head-training & 0.119 & 0.321 & 0.201 & \textbf{0.356} \\
\bottomrule
\end{tabular}
}
\end{subtable}

\vspace{1em}
\begin{subtable}{\linewidth}
\centering
\caption{Source: Caltech101}
\label{tab:caltech101}
\resizebox{0.8\linewidth}{!}{
\begin{tabular}{lccc|c}
\toprule
Score & ETran & SFDA & LogME & Wasserstein \\
\midrule
Head-training & 0.290 & 0.301 & 0.198 & \textbf{0.325} \\
\bottomrule
\end{tabular}
}
\end{subtable}

\vspace{1em}
\begin{subtable}{\linewidth}
\centering
\caption{Source: Caltech256}
\label{tab:caltech256}
\resizebox{0.8\linewidth}{!}{
\begin{tabular}{lccc|c}
\toprule
Score & ETran & SFDA & LogME & Wasserstein \\
\midrule
Head-training & \textbf{0.431} & 0.384 & 0.295 & 0.363 \\
\bottomrule
\end{tabular}
}
\end{subtable}

\vspace{1em}
\begin{subtable}{\linewidth}
\centering
\caption{Source: Flowers102}
\label{tab:flowers102}
\resizebox{0.8\linewidth}{!}{
\begin{tabular}{lccc|c}
\toprule
Score & ETran & SFDA & LogME & Wasserstein \\
\midrule
Head-training & 0.305 & \textbf{0.412} & 0.384 & 0.369 \\
\bottomrule
\end{tabular}
}
\end{subtable}

\end{table}

\begin{table}[t]
\centering
\caption{Aggregated results of Table \ref{tab:fine_tuning} showing our proposed metric have a better performance for head-training: This table presents the average correlation (weighted Kendall's tau) across five different source datasets (ImageNet, CIFAR100, Caltech101, Caltech256, and Flowers102).}
\label{tab:aggregated_results}
\resizebox{0.8\linewidth}{!}{
\begin{tabular}{lccc|c}
\toprule
& ETran & SFDA & LogME & Wasserstein \\
\midrule
Head-training & 0.315 & 0.374 & 0.316 & \textbf{0.387} \\
\bottomrule
\end{tabular}
}
\end{table}

Head-training is the method that keeps the feature extractor layers of the source model fixed and then trains a new head classifier using the target dataset before re-training the new classifier~\cite{Nguyen2020}.

Table~\ref{tab:aggregated_results} shows that our weight-based metrics, particularly the Wasserstein distance, perform consistently well regardless of the fine-tuning strategy employed. Notably, our weight-based metric improves performance in the head-training strategy by a relative \textbf{+3.5\%} compared to the next best method (0.387 vs. 0.374 for SFDA~\cite{Shao2022}). This contrasts with existing methods such as LogME~\cite{You2021}, and ETran~\cite{Gholami2023}. The robustness of our weight-based approaches across different transfer learning scenarios is a significant advantage, allowing practitioners to reliably select models regardless of their intended fine-tuning approach.

\subsubsection{Computational Time}

Table~\ref{tab:computational_time} presents the computational time required by our proposed method for fine-tuning on pseudo labels across five different datasets. The results show the time requirements for both head-training and full-training strategies. Head-training computational times range from 144 seconds for the Aircraft dataset to 81 seconds for the DTD dataset. Full-training times vary from 425 seconds for Aircraft to 278 seconds for DTD. We observe that computational requirements generally correlate with dataset complexity, with larger datasets requiring more processing time. It is worth noting that these results were obtained using only two epochs of fine-tuning; additional epochs would likely yield improved performance at the cost of longer computational times. However, even with just two epochs, our method achieves competitive results while maintaining computational efficiency, making it particularly suitable for realistic scenarios.

\begin{table}[t]
\centering
\caption{Computational time (in seconds) required for fine-tuning on pseudo labels for two epochs using our proposed method across different datasets.}
\label{tab:computational_time}
\resizebox{\linewidth}{!}{
\begin{tabular}{lccccc}
\toprule
Fine-tuning Strategy & \multicolumn{5}{c}{Dataset} \\
\cmidrule{2-6}
& Aircraft & Caltech101 & Flowers102 & Pets & DTD \\
\midrule
Head-training & 144s & 129s & 114s & 98s & 81s \\
Full-training & 425s & 395s & 358s & 312s & 278s \\
\bottomrule
\end{tabular}
}
\end{table}

\section{Conclusion}

In this paper, we introduced TransferTest, a comprehensive platform for unbiased evaluation of transferability estimation methods. Through extensive experiments across different source datasets, model complexities, and fine-tuning strategies, we demonstrated that:

\begin{enumerate}
\item Traditional transferability metrics heavily rely on ImageNet pre-trained models and may not generalize well to other pre-training datasets.

\item Our proposed weight-based metrics, particularly the Wasserstein distance between original and fine-tuned model weights, provide robust transferability estimates across various conditions.

\item The effectiveness of transferability estimation methods varies with model complexity, but weight-based metrics show greater stability across different model subsets.

\item The choice of fine-tuning strategy significantly impacts some transferability metrics, while our weight-based approaches remain consistently effective. Notably, our proposed Wasserstein metric achieves a 3.5\% relative improvement over the next best method in head-training scenarios, demonstrating its superior performance in this important practical setting.

\end{enumerate}

REPT provides a structured framework for evaluating transferability estimation methods under diverse, realistic scenarios, addressing the limitations of previous evaluation approaches that focused primarily on ImageNet-trained models and full fine-tuning. Our results suggest that weight-based transferability metrics, particularly using Wasserstein distance, offer a promising direction for practical transferability estimation in real-world transfer learning applications.

Future work could explore the combination of feature-based and weight-based approaches. Additionally, investigating the theoretical connections between weight distributions and transferability could provide deeper insights into the transfer learning process.
\\

{
    \small
    \bibliographystyle{ieeenat_fullname}
    \bibliography{main}  
}

\end{document}